\documentclass[10pt, conference, compsocconf]{IEEEtran}
\ifCLASSINFOpdf
\else
\fi
\usepackage{graphicx}
\usepackage{float} 
\usepackage{multirow}
\usepackage{microtype} 
\usepackage{placeins} 
\usepackage{color}
\usepackage{textcomp}

\usepackage{fancyhdr,lipsum}
\fancypagestyle{firstpage}{% Page style for first page
  \fancyhf{}% Clear header/footer
  
  \fancyhead[C]{2017 14th Conference on Computer and Robot Vision}% Header
  \fancyfoot[C]{\thepage}% Footer
  \fancyfoot[L]{978-1-5386-2818-8/17 \$31.00\ \textcopyright\ 2017 IEEE \\ DOI 10.1109/CRV.2017.42}
}
\begin{document}
\setcounter{page}{232}

%\title{Learning Robust Object Recognition with Imagination}
%\title{Learning Robust Object Recognition by Training Networks with Composed Scenes Using Generative Models}
\title{Learning Robust Object Recognition Using Composed Scenes \\from Generative Models}


\author{\IEEEauthorblockN{Hao Wang, Xingyu Lin}
\IEEEauthorblockA{Department of Computer Science\\
Peking University\\
Beijing, China\\
\{hao.wang, sean.linxingyu\}@pku.edu.cn}
\and
\IEEEauthorblockN{Yimeng Zhang, Tai Sing Lee}
\IEEEauthorblockA{Department of Computer Science\\
Carnegie Mellon University\\
Pittsburgh, USA\\
yimengzh@cs.cmu.edu, tai@cnbc.cmu.edu}
}

\maketitle
\thispagestyle{firstpage}
\lipsum[232-239]

\begin{abstract}
Recurrent feedback connections in the mammalian visual system have been hypothesized to play a role in synthesizing input in the theoretical framework of analysis by synthesis. The comparison of internally synthesized representation with that of the input provides a validation mechanism during perceptual inference and learning. Inspired by these ideas, we proposed that the synthesis machinery can compose new, unobserved images by imagination to train the network itself so as to increase the robustness of the system in novel scenarios. As a proof of concept, we investigated whether images composed by imagination could help an object recognition system to deal with occlusion, which is challenging for the current state-of-the-art deep convolutional neural networks. We fine-tuned a network on images containing objects in various occlusion scenarios, that are imagined or self-generated through a deep generator network. Trained on imagined occluded scenarios under the object persistence constraint, our network discovered more subtle and localized image features that were neglected by the original network for object classification, obtaining better separability of different object classes in the feature space. This leads to significant improvement of object recognition under occlusion for our network relative to the original network trained only on un-occluded images. In addition to providing practical benefits in object recognition under occlusion, this work demonstrates the use of self-generated composition of visual scenes through the synthesis loop, combined with the object persistence constraint, can provide opportunities for neural networks to discover new relevant patterns in the data, and become more flexible in dealing with novel situations. 
\end{abstract}


% \fi
\IEEEpeerreviewmaketitle

\section{Introduction}
% Background
It is well known that there is an enormous amount of recurrent feedback connections in visual systems, yet their functions remain elusive. One conjecture is that they can help performing ``analysis by synthesis''---via priors encoded in the recurrent connections, higher visual areas synthesize representations in the lower ones to explain the input stimuli \cite{Mumford:1992cw}, and this strategy is important for perceptual inference and learning. In the mammalian memory system, there is synthesis in the form of reactivation or ``replay'' of experienced events during sleep \cite{Wilson:1994vw}, and such ``replay'' is critical to memory consolidation \cite{McClelland:1995jp}. In computational models such as Boltzmann machine \cite{Hinton:1984us} and deep belief nets \cite{Hinton:2006kc}, learning is done by matching the statistics of the training data and those of the model generated through some synthesis processes (such as sampling).

% Background
In addition to replaying or matching experienced events, the synthesis in the brain exhibits much diversity: the brain may play memory forward, backward, or in new combinations, reflecting introspection and imagination \cite{Gupta:2010go,Derdikman:2010cw}. The composition of new, unobserved scenes and events can be due to random co-activations of replay processes through the recurrent interaction between hippocampus and the neocortex, and/or the recurrent loops in the visual system.

% Method
How could this imagination process be useful for learning? We argue that a system may create new ``experienced'' events and imagine scenarios via already learned rules and constraints about the environment, such as spatial relationships among objects, to train itself for unobserved novel tasks and situations. This idea seems paradoxical at first: what new things can a system learn from what it already knows? On the other hand, as many artists may testify, we can often learn and discover new insights by looking at our own creative composition. In this paper, we explored this idea in the context of object recognition under occlusion. In particular, we generated an ``imagined'' data set by combining labeled images from ImageNet \cite{Deng:2009dl} with objects rendered or generated by generative neural networks to create new composition of scenes with different types and levels of occlusion, and fine-tuned a deep convolutional neural network (DCNN) on this data set under the object persistence principle. The object persistence principle states that after we see an object, we believe it continues to exist within a short-time window, and the neurons coding for the objects will continue to fire even when the object translates, or rotates in 3D space or simply disappears behind a certain occluder. This assumption about temporal persistence of objects has been exploited in slow-feature analysis \cite{Wiskott:2002ue} or memory trace models \cite{Perry:2006gl} in computational neuroscience to learn translation and rotation invariant object recognition. Here, we extended this idea by assuming that once an object is seen, the object labeling neurons will continue to fire even when the object is occluded. This principle was implemented by clamping the teaching signals of the original labelled images under various occlusion conditions in the composed images. 

% Result
Through this ``training by imagining'' procedure, the fine-tuned network discovered more subtle and localized image features important for classifying different objects, and achieved significantly better performance under various occlusion scenarios with little performance compromise, compared to our baseline model trained with data from ImageNet only. This suggested that the synthesis process potentially available in the brain could be useful for teaching the system to perform better under novel and unobserved situations.

\begin{figure}[H]
    \centering
    \includegraphics[width=1.0\columnwidth]{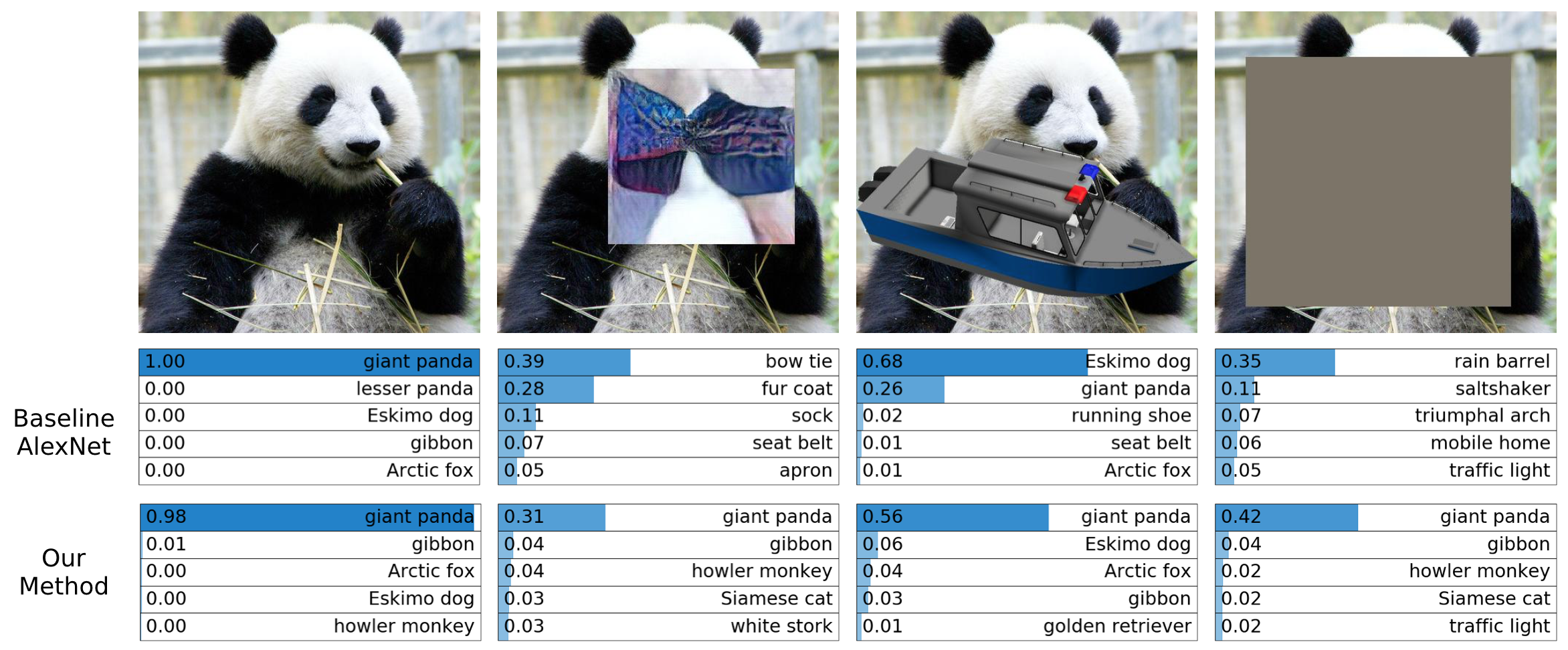}
    \caption{Different types of occlusion and their effects on the output (top-5 confidence) of the networks. First row: the original image and its occluded versions. Second: output of an AlexNet fine-tuned on 100 classes' images without occlusion (``baseline'' in our experiments). Third: output of the network trained using our method (``imagination'' in our experiments).}
    \label{fig:occlusion_examples}
\end{figure}

\section{Related Work}
As occlusion is a crucial issue in various computer vision problems, it has been studied for more than a decade. The methods developed to addressing occlusion problems can be categorized into three types: domain-specific methods, part-based models and neuro-inspired approaches.

\subsection{Domain-specific Methods}
Generally, domain-specific methods deal with occlusion of a certain kind of object by acquiring prior knowledge from people. For pedestrian detection, \cite{Wu:2005fm} employs a Boltzmann distribution to learn the prior of local deformation. Occlusion is also a challenging problem in face recognition. \cite{Rama:2008ks} compares three different PCA strategies in robust face recognition, including holistic PCA, component-bases PCA and near-holistic PCA. While domain-specific methods work well in certain subfields of computer vision, it's hard to apply these ad hoc solutions to problems in other domains, let alone to transfer models' ability learned from one domain to another.

\subsection{Part-based Models}
As occlusion normally does not block every distinct components of objects, part-based models try to utilize this nature by considering objects as a composition of several parts. Furthermore, part-based models usually require prior knowledge of the set of parts for each class. Deformable part model \cite{Felzenszwalb:2010ez} is a successful approach in this area, inspired by which \cite{Ghiasi:2014je} proposes a hierarchical DPM to localize occluded faces. Although the part-based models succeed in certain tasks, they have some drawbacks. In addition to that these models may require data sets with hand-annotated parts to handle occlusion, there is actually a considerable number of objects not easily decomposable into distinct parts, such as flexible animals like snakes or objects without parts like potatoes. Thus, the part-based models are not suitable for all kinds of objects and lack scalability. 

\subsection{Neuro-inspired Approaches}
Early works of neuro-inspired approaches are basically based on shallow neural network models. \cite{Fukushima:2000jr} utilizes neural network model to recognize occluded patterns. \cite{Fukushima:2005gx} restores occluded parts of a pattern by feedback signals from the top stage of the network. However, these initial works use primitive models with low capacity and can only be applied to toy data sets, such as simple letter patterns. In recent years, deep convolutional neural networks represented by AlexNet \cite{Krizhevsky:2012wl} have shown strong ability on various kinds of large-scale computer vision tasks, including object detection, localization and classification. By designing deep architectures and taking advantage of high-performance computing resources, they have been successful on large scale image databases including ImageNet \cite{Deng:2009dl}. Based on recent progress on deep networks, our approach is a simple yet powerful way to handle occlusion problems as well as provides generalisation ability and scalability.

\section{Methods}
\begin{figure}[H]
    \centering
    \includegraphics[width=1.0\columnwidth]{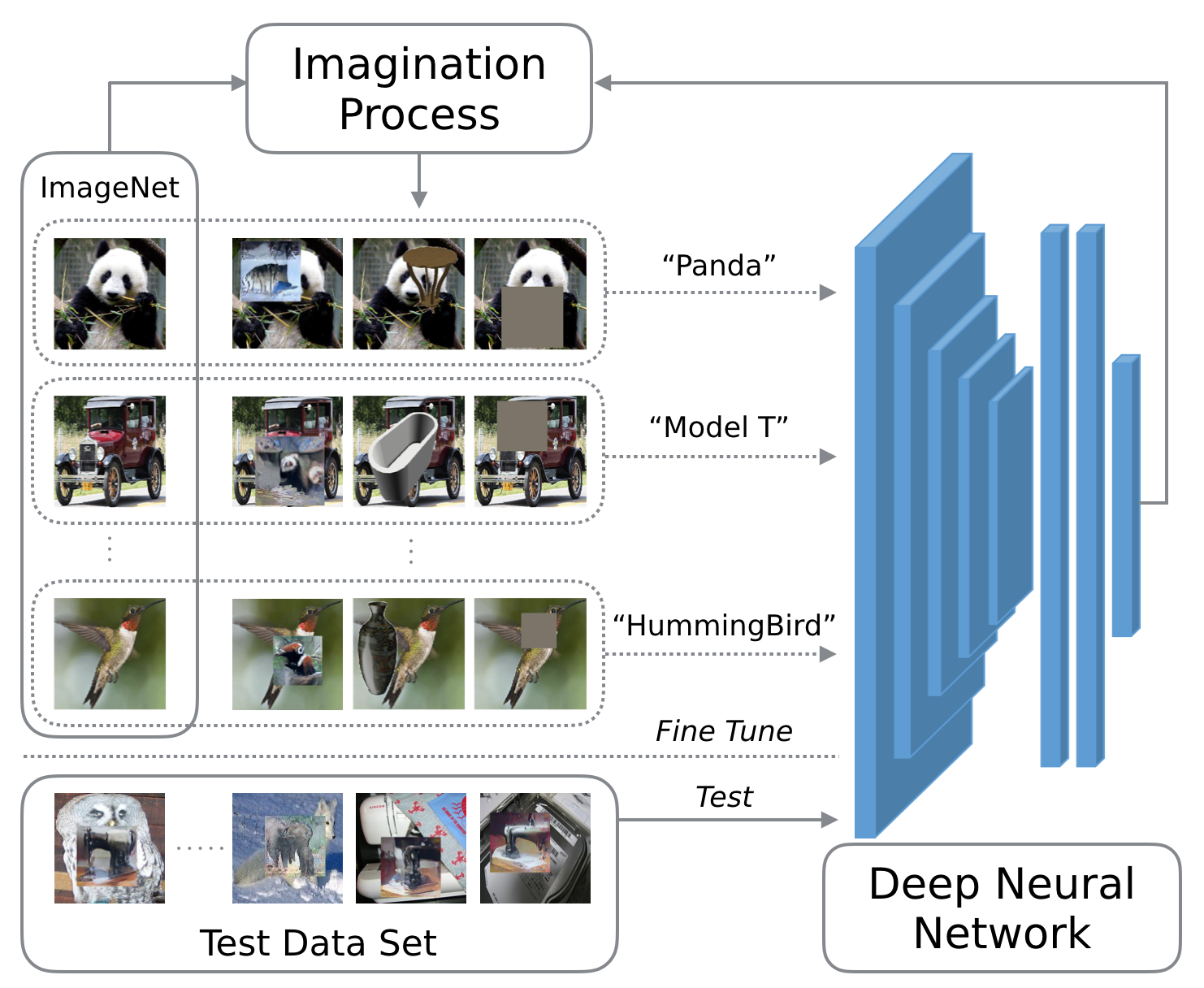}
    \caption{The framework of our approach.}
    \label{fig:framework}
\end{figure}

% Description, overall
Deep convolutional neural network (DCNN), apart from performing better than other computer vision models in general, is also a reasonable model for the feedforward computation in the visual cortex. Thus, we implemented our ``training by imagining'' procedure based on a standard DCNN architecture (AlexNet). While AlexNet can perform standard object recognition well by training on ImageNet \cite{Deng:2009dl}, their performance would drop significantly when objects are occluded significantly (Figs.~\ref{fig:occlusion_examples}, \ref{fig:accuracy}). We investigated whether a DCNN can be trained to deal with occlusion better using synthesized images composed of the original labeled images with synthesized occluded objects added. These training images are synthesized either manually to simulate visual experience of object persistence, or they can be internally generated by the network itself to simulate the imagination process (Fig.~\ref{fig:framework}). The DCNN is pre-trained with images without occlusion to learn features of each class. In the fine-tuning, first an ImageNet image is presented with the corresponding object unit at the top layer fixed to be active (``panda'' and ``model T'' in Fig.~\ref{fig:framework}). Then, an occluding object is synthesized during the imagination process and added to the image to occlude the target object in the original scene. In these synthesized images, objects can occlude each other in various ways. They are then used to train the network with the original target unit clamped active. Thus, the network is instructed to believe the target object is in the scene regardless of the level of occlusion.

% Method: imagination process
There are many candidate algorithms for implementing the imagination process to generate the imagined occluded images, such as algorithms for AND/OR graph \cite{Zhu:2006bu}, deep belief nets \cite{Hinton:2006kc}, deconvolutional networks \cite{Zeiler:2013ws}, Google DeepDream, and adversarial networks \cite{Goodfellow:2014td,Wang:2016tt}. Here we concatenated the deep generator network (DGN) proposed in \cite{Nguyen:2016ub} with a network trained on data sets without occlusion used as the target DNN to implement the imagination process. Under this framework, an occluder image is synthesized during an activation maximization (AM) process. The AM process maximize activation of a particular neuron in the target DNN's top fully connected layer through standard backpropagation method, which only changes the DGN input code rather than weights of the DGN and the target DNN (more detailed description of DGN and AM can be found in \cite{Nguyen:2016ub}). Thanks to the natural image priors provided in DGN, the synthesized occluders can be quite realistic-looking. We used different target DNNs to synthesized various types of occlusion (Fig.\ref{fig:occluder}) and added them to the original image in a composition process. The details of generation of each type of occlusion will be discussed in the experiments section.

% Test & Analyze
After training the network using synthesized images, novel objects under a variety of occlusion scenarios were used to test the system's robustness in object recognition against occlusion. In addition, we analyzed the feature representation of the networks before and after the fine-tuning, trying to understand what changes had taken place in these representations and why these changes might allow the networks to deal with the novel situations of occlusion.

% Obstacles
However, two major potential conceptual obstacles exist in our approach. First, when an object of class B occludes the target object of class A in an imagined image, if the network is instructed, via a supervision signal due to the object persistence constraint, to consider the image as an instance of A, how can we prevent the system from learning to classify object B as object A? Second, when the target object A is completely occluded or becomes invisible in complete darkness, what will prevent complete darkness from being learned to be recognized as object A --- that is, what will prevent the network from thinking that object A is there even when none of its parts is shown? As it turned out, the first problem is immaterial when a sufficient variety of objects are used for occlusion in the training set, i.e., the features from different occluding objects will be averaged out over time, leaving only the critical and persistent features of the target object to be learned for classification. The second problem is solved when many classes of objects are trained together. While it is true that complete darkness or occlusion might be falsely associated with target object A, the situation is also true for object B and object C when they are being trained as targets. Thus, complete occlusion or darkness will be associated with all classes of objects with uniform and tiny probabilities.
\section{Experiments}
\subsection{Experiment Settings}
\subsubsection{Data sets}
We randomly sampled 100 classes from ImageNet, and extracted 500 images with labeled bounding boxes from each class. We then partitioned all images randomly into training data, validation data, and test data with a ratio of 3:1:1. With roughly the same number of images of each class for every data set, we addressed the second issue discussed in the Methods section. In order to eliminate the interference of background information, we cropped all images so that only pixels in the bounding boxes were left.

We derived two sub training data sets from the training data. The first one was the original data set without the imagination process. The network fine-tuned with this data set is called ``baseline''. The second one was a data set generated with our imagination process, containing images with different occlusion types (including ImageNet object, ShapeNet object and gray rectangular occlusion, see below) and occlusion levels (0\%, 10\%, 20\%, ..., 100\%), and image labels were kept unchanged. We call the network fine-tuned with this ``imagined'' data set ``imagination''. Similarly, we prepared original and imagined versions for validation and test data sets.

\subsubsection{Configurations of occlusion}
We used several types of occlusion for the imagination process, including ImageNet object occlusion, ShapeNet object occlusion and gray rectangular occlusion (Fig.~\ref{fig:occlusion_examples}). For any image, we put one occlusion of certain type and level within each of its bounding boxes containing the labeled objects.

\begin{figure}[H]
    \centering
    \includegraphics[width=0.8\columnwidth]{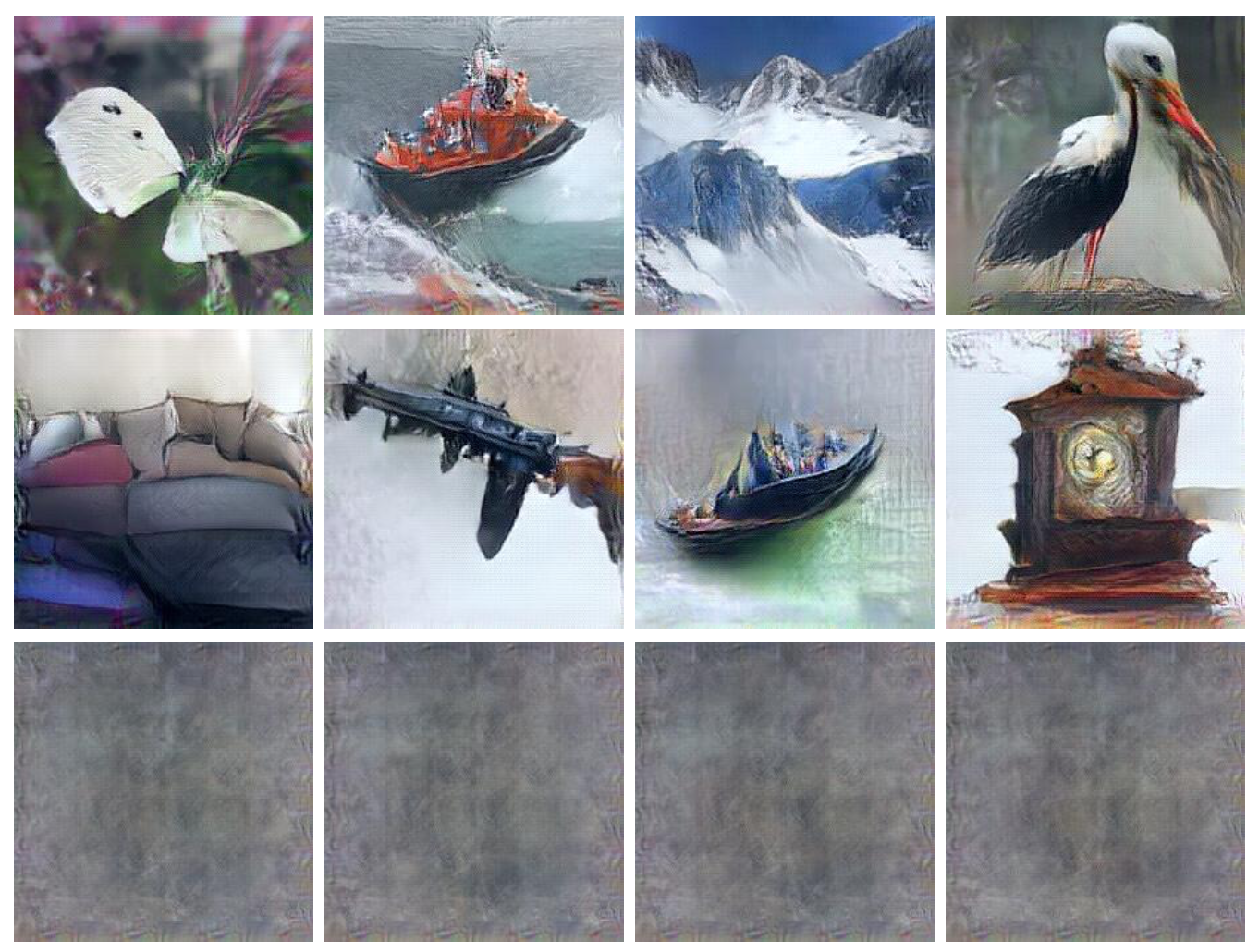}
    \caption{Different types of synthesized occluders. First row, example occluders synthesized from a network fine-tuned with ImageNet data. Second row, example occluders synthesized from a network fine-tuned with ShapeNet data. Third row, example occluders of gray rectangular occlusion (minor difference).}
    \label{fig:occluder}
\end{figure}

\paragraph{ImageNet Object Occlusion}
The ImageNet object occlusion scenes (Fig.~\ref{fig:occlusion_examples}, second column) are created by superimposing the generated images as occluders with the original labeled images. In order to generate these occluders, we assigned the ``baseline'' network as the target DNN in the DGN framework. For each class among the sampled 100 classes, we maximize the activation of the corresponding neuron in the ``baseline'' network's top fully connected layer. 20 occluder images from each class were generated (Fig.~\ref{fig:occluder}, first row). The generated occluders were scaled and randomly placed on the original image by specifying a particular area in the image it would go. The occlusion level is defined by the ratio of the area of the generated image over the area of the bounding box of the targeted object in the original image. It is worth mentioning that, the scaling and positioning of the occluder object image can also be accomplished automatically in the synthesis process by clamping the pixels in the imagined image outside the targeted area to zero, which will be treated transparent when being superimposed on to the original labeled image during the composition process. 

\paragraph{ShapeNet Object Occlusion}
While the ImageNet object occlusion introduces occluders within the target classes to be classified, we also consider a type of occlusion, the ShapeNet \cite{Chang:2015uq} object occlusion (Fig.~\ref{fig:occlusion_examples}, third column), which employs occluders that the network won't classify. This is similar to human visual experience, in which people might see occluders that they don't yet recognize. Similar to the ImageNet object occlusion, one way to deploy the ShapeNet object occlusion is to synthesize occluders by using DGN and a network trained with ShapeNet data (Fig.~\ref{fig:occluder}, second row). Here, for simplicity and efficiency, we alternatively use ShapeNet data directly. We first randomly sample one from the rendered images of 1000 objects from ShapeNet. We then overlay the rendered image onto each bounding box of the occluded image in random position. The occlusion level is defined as the ratio of the size of the occluding object's bounding box over that of the occluded object's bounding box. However, aspect ratios of these two bounding boxes are usually different. In order to maintain the aspect ratio of the occluding object as much as possible, we adopted a strategy that, while keeping the aspect ratio of the occluding object when the occlusion level is low, we stretch the occluding object a little when the occlusion level is too large to fit the occluding object into the bounding box of the occluded objects.

\paragraph{Gray Rectangular Occlusion}
The rectangular occlusion (Fig.~\ref{fig:occlusion_examples}, fourth column) is defined as a gray rectangle within the bounding box. In the framework of our approach, by calculating the mean image of images synthesized from each class, we can synthesize a gray rectangle occluder (Fig.~\ref{fig:occluder}, third row). Here we use an alternative approach, generating rectangles with mean color of the ILSVRC 2012 training set, which has purer color but identical function as the synthesized one. Different from the first two types of occlusion, the gray rectangular occlusion cause incomplete information instead of introducing interference information. The rectangle has the same aspect ratio as the bounding box, as a result of which the rectangle will actually be a square when rescaled and sent as an input to AlexNet. The occlusion level is given by the ratio of the area of the rectangle over that of the bounding box.

\subsubsection{Parameter settings for fine tuning}
We used Caffe \cite{Jia:2014up} to perform experiments. We initialized the weights by a pre-trained AlexNet network and then fine-tuned the top two fully connected layers. The fine-tuning procedure was accomplished through minimizing the multinomial logistic loss by back-propagation \cite{LeCun:1989bx}. We used mini-batch gradient descent method with momentum, in which the batch size was 256 and the momentum was 0.9. The fine tuning process started from an initial learning rate of 0.0005, which decreased by a factor of 10 for every 5000 iterations, and stopped after maximally 10000 iterations. The weight decay was set to 0.001.

\subsection{Experiment Results}
% [Overall Accuracy]
\subsubsection{Overall accuracy}

\begin{figure}[H]
    \centering
    \includegraphics[width=\columnwidth]{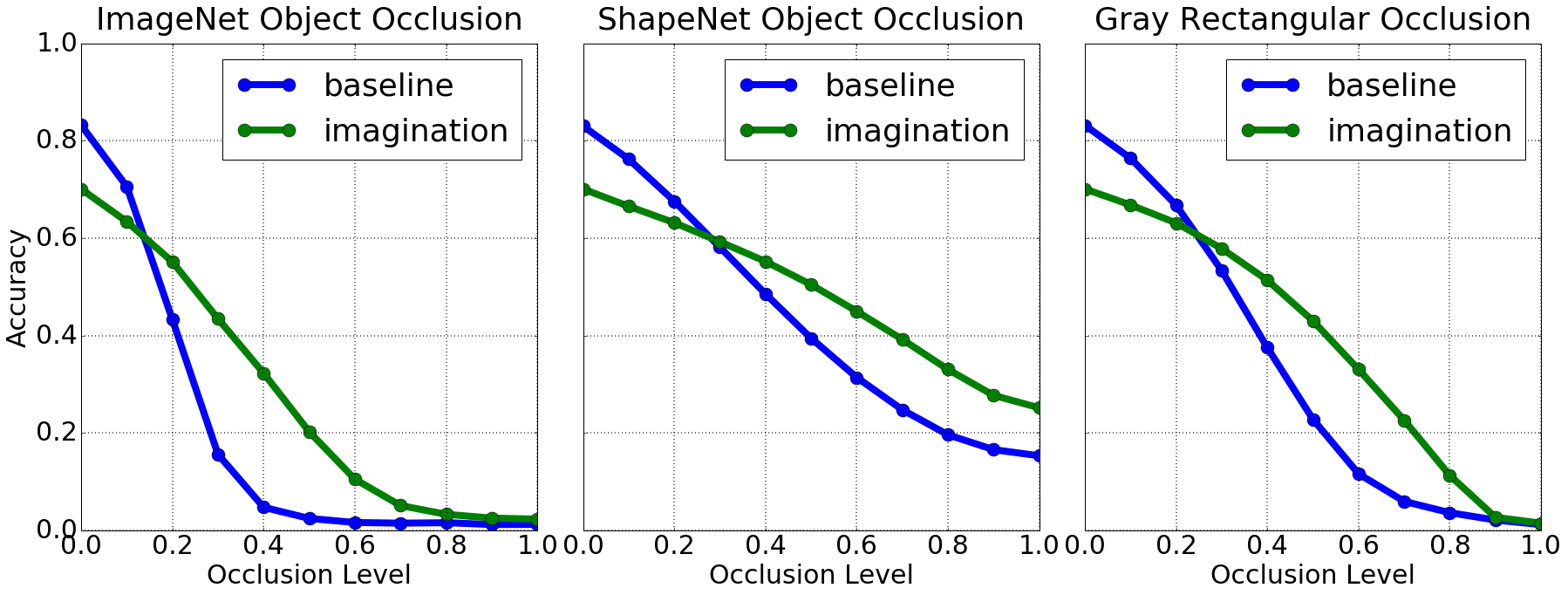}
    \caption{Test accuracy on different types and different levels of occlusion. Each panel shows the result for one type of occlusion. }
    \label{fig:accuracy}
\end{figure}

By performing fine-tuning on either of original or ``imagined'' data sets, we obtained two networks ``baseline'' and ``imagination''. Based on test results on validation sets, these two networks converged without overfitting or underfitting. Fig.~\ref{fig:accuracy} and Tbl.~\ref{table:accuracy} show their performance on test data sets of different occlusion types and levels.

For the test data sets with ImageNet object occlusion, the ``imagination'' network significantly outperformed the ``baseline'' network in overall top-1 classification accuracy with only a little compromise on classifying images without occlusion. When tested on the data set with 30\% ImageNet object occlusion, the ``imagination'' network even gained an increase of 27.9\% in accuracy over the ``baseline'' network. Similar results were observed on other test data sets. Given that we only used one network and achieved significantly better performance over the baseline model under different types and levels of occlusion, our ``training by imagining'' approach seems to show surprising generality and robustness to various occlusion scenarios.

\begin{table}[H]
\centering
\resizebox{\columnwidth}{!}{\begin{tabular}{|c|c|c|c|c|c|c|c|c|c|c|c|}
\hline
Occlusion Level & 0\% & 10\% & 20\% & 30\% & 40\% & 50\% & 60\% & 70\% & 80\% & 90\% & 100\%  \\ \hline
Baseline      & 0.832 & 0.706 & 0.433 & 0.156 & 0.047 & 0.024 & 0.016 & 0.015 & 0.016 & 0.012 & 0.012 \\ \hline
Imagination   & 0.701 & 0.634 & 0.551 & 0.434 & 0.323 & 0.202 & 0.105 & 0.051 & 0.033 & 0.025 & 0.023 \\ \hline
Increase      & -0.131 & -0.072 & 0.119 & \bf{0.279} & 0.276 & 0.178 & 0.089 & 0.036 & 0.017 & 0.013 & 0.011 \\ \hline
\end{tabular}}
\caption{Test accuracy on different levels of ImageNet object occlusion. The accuracy increase achieved its maximum at the occlusion level of 30\%.}
\label{table:accuracy}
\end{table}

% [Accuracy Increase for Each Class]
\subsubsection{Accuracy improvement for each class}
\begin{figure}[H]
    \centering
    \includegraphics[width=0.8\columnwidth]{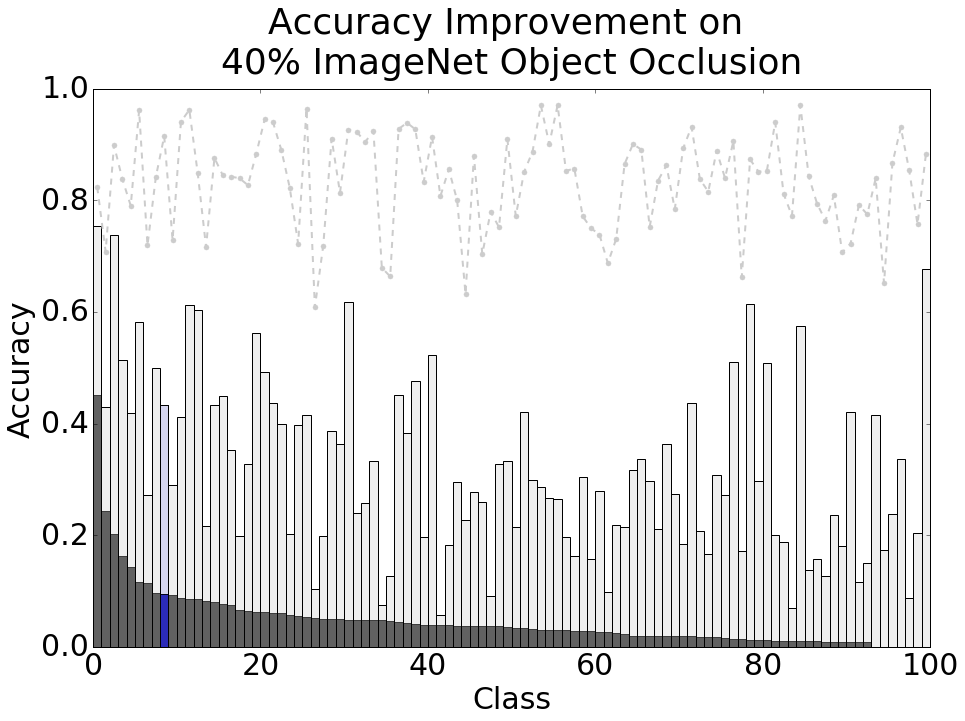}
    \caption{Test accuracy improvement on 40\% ImageNet object occlusion. The dark gray bars represent accuracy of the ``baseline'' network for each class, by which the bars are sorted. The light gray bars stand for accuracy increase of the ``imagination' network w.r.t. the ``baseline'' network for each class. The light gray line shows the test result of ``baseline'' network on images without occlusion, which serves as an upper bound for possible accuracy. The highlighted blue bar indicates the ``panda'' class mentioned in Methods section.}
    \label{fig:accruacy_improvement}
\end{figure}

We further investigated the performance improvement for each class. Taking the 40\% ImageNet object occlusion test data set as an example, we plotted the accuracy of the ``baseline'' network as well as the accuracy increase of the ``imagination'' network for each class (Fig.~\ref{fig:accruacy_improvement}). It turned out that, with the overall accuracy increase of 27.6\%, most classes gained considerable accuracy improvements, some of which even had increases around 60\%. As a side-point, when entire images (instead of images containing only bounding boxes) were used to train an ``imagination'' network, it could get an accuracy of 27.5\% on the ``imagined'' full-image test data set, even when the objects were completely occluded in test data sets. This was because the background context is often correlated with the object in ImageNet.

\subsection{Analysis}
As we showed that our ``training by imagining'' approach which utilizes the concept of object persistence significantly improved the network's robustness to occlusion, a natural question arises: why does it work? In this section, we attempt to answer this by performing analysis from several perspectives.

\subsubsection{Learning from different levels and different types of occlusion}
\begin{figure}[htbp]
    \centering
    \includegraphics[width=1.0\columnwidth]{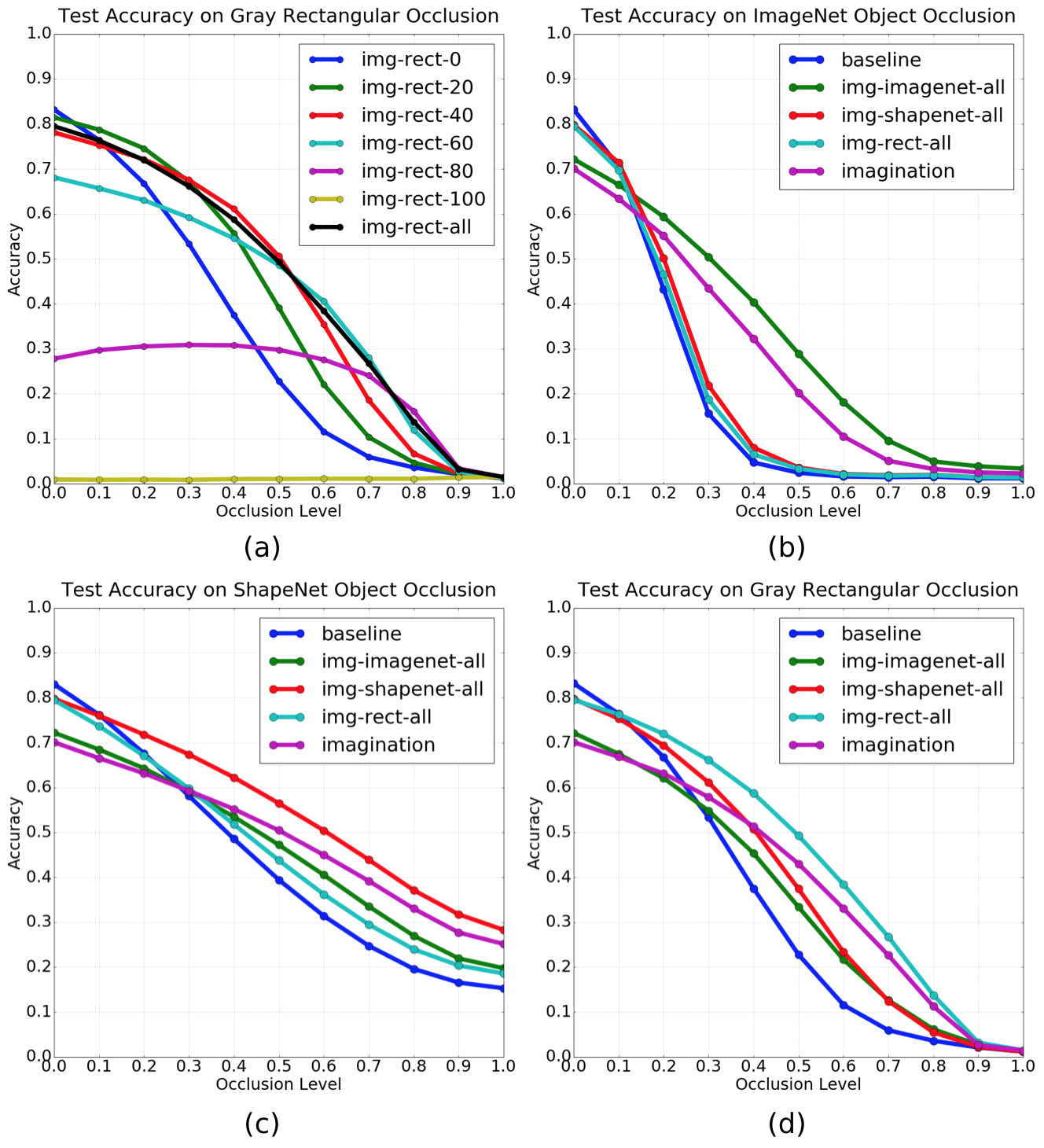}
    \caption{(a) Test accuracy on different levels of gray rectangular occlusion. Each line represents an accuracy curve of a network fine-tuned with certain level of rectangular occlusion. (b)(c)(d) Test accuracy on three types of occlusion. }
    \label{fig:analysis_level_type}
\end{figure}

First of all, the consistent performance improvement of the ``imagination'' network was partially due to the variety of occlusion types and levels in the training data set. To show this, we broke down the imagination process by the level and the type of occlusion. For the analysis on the effect of different occlusion levels, we take the gray rectangular occlusion as an example and trained several networks. Each of them was fine-tuned with a training data set with images with a particular level ($x\%$) of gray rectangular occlusion, named ``img-rect-$x$'', while the "img-rect-all" was a network fine-tuned with all levels of gray rectangular occlusion. Then we tested these networks on data sets of different levels of gray rectangular occlusion (Fig.~\ref{fig:analysis_level_type}(a)). For the analysis on the effect of different occlusion types, we also trained several networks, each of which was fine-tuned with a training set with images of occlusions of type $T$ in all levels, named ``img-$T$-all''. We then tested them along with the ``imagination'' network on data sets of different types and levels of occlusion (Fig.~\ref{fig:analysis_level_type}(b)(c)(d)).

The analysis on different occlusion levels (Fig.~\ref{fig:analysis_level_type}(a)) showed that the ``img-rect-$x$'' network always performed better than other models on the test data set with $x\%$ occlusion. Meanwhile, it performed worse than others in other levels of occlusion. The analysis on different occlusion types (Fig.~\ref{fig:analysis_level_type}(b)(c)(d)) showed that the ``img-$T$-all'' network always performed best on the test data set with occlusion type $T$. Combining these two observations, we conclude that one reason why the ``imagination'' network performs well is that, it handles certain level and type of occlusion by learning from it. Furthermore, by analyzing the accuracy of the ``img-$T$-all'' networks across different types of occlusion, we find there exists a certain level of transferability. For example, the ``img-imagenet-all'' network transfers to ShapeNet Object occlusion, and the ``img-shapenet-all'' network transfers to gray rectangular occlusion. This kind of transferability hints that, the occlusion a network fine-tunes on and the occlusion a network transfers to may have some shared attributes. Finally, as expected, the ``imagination'' network gained the best transferability and the highest overall accuracy.

In short, by learning from images of different levels and types of occlusion, the ``imagination'' network gained overall highest performance under various occlusion scenarios.

\subsubsection{Learning Occlusion Invariance Features}
\begin{figure*}[htbp]
    \centering
    \includegraphics[width=0.8\textwidth]{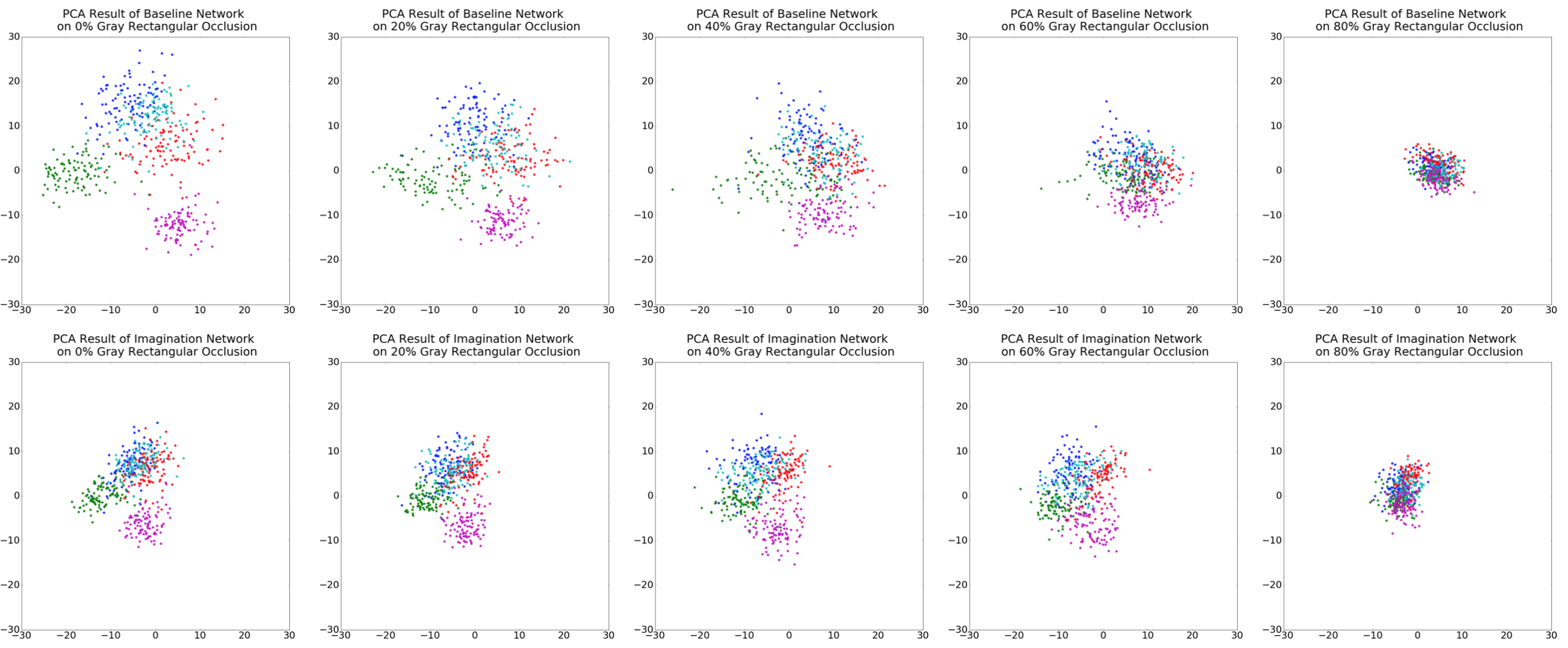}
    \caption{PCA result of the ``baseline'' network and the ``imaginaton'' network on different levels of gray rectangular occlusion. Each color represents a class. We use same projection for all feature vectors in the figure.}
    \label{fig:analysis_PCA}
\end{figure*}

Second, ``imagination'' network learns occlusion invariance features of its top layer output, allowing for easier separation and clustering of different object classes under high level of occlusion. To quantify the difference of feature space separability between ``baseline'' and ``imagination'' networks, here we define a score function $J(\cdot)$ for measuring the ratio of inter-class distance over intra-class distance for top layer feature vectors (fc8) of networks, inspired by the Linear Discriminant Analysis (LDA) \cite{Rao:1948iw}:
\begin{equation}
J(S)=\frac{d_{inter}}{d_{intra}}=\frac{\frac{1}{|S|}\sum_{c}||\mathbf{\mu_{c}}-\mathbf{\bar{x}}||^{2}}{\frac{1}{|S|}\sum_{c \in S}\frac{1}{|c|}\sum_{\mathbf{x} \in c}||\mathbf{x}-\mathbf{\mu_{c}}||^{2}},
\end{equation}
where $S$ is the set of 100 classes, $c$ is a feature vector set of one class, $\mathbf{\mu_{c}}$ is the mean of feature vectors in one class, $\mathbf{\bar{x}}$ is the mean of feature vectors in all classes and $\mathbf{x}$ is a single feature vector in one class. In this score function, the numerator is the variance of class centroids, representing the average inter-class distance, and the denominator is the mean of variance of feature vectors in each class, standing for the average intra-class distance. Higher $J$ means better separability between different classes.

We calculated inter-class distance $d_{inter}$, intra-class distance $d_{intra}$, and the score $J$ for the ``baseline'' network and the ``imagination'' network on the test data sets with different levels of ImageNet object occlusion (Tbl.~\ref{table:LDA}). For both networks, with higher occlusion levels, the inter- and intra-class distances among distinct objects started to collapse as more and more image features originally associated with the objects got occluded. However, training with the imagined data somehow reversed this trend, as shown by relatively higher $J$ of the ``imagination'' network over the ``baseline''. The result showed the direct reason why the network performed well: the network learns occlusion invariance features by being fine-tuned with the imagination process, with overall better separability of objects of different classes (higher $J$) at various occlusion levels. Similar results were obtained on other occlusion types. 

\begin{table}[ht]
\centering
\resizebox{\columnwidth}{!}{\begin{tabular}{|c|c|c|c|c|c|c|c|c|c|c|c|c|}
\hline
& Occlusion Level & 0    & 10   & 20   & 30   & 40   & 50   & 60   & 70   & 80   & 90   & 100  \\ \hline
\multirow{2}{*}{$J$} & Baseline     & 1.58 & 1.25 & 1.02 & 0.79 & 0.58 & 0.40 & 0.28 & 0.20 & 0.12 & 0.12 & 0.37 \\
& Imagination  & 0.87 & 0.80 & 0.78 & 0.75 & \bf{0.70} & \bf{0.61} & \bf{0.50} & \bf{0.38} & \bf{0.24} & \bf{0.15} & 0.42 \\ \hline
\multirow{2}{*}{$d_{inter}$} & Baseline & 497.3 & 360.2 & 276.3 & 197.8 & 126.4 & 68.9 & 33.7 & 15.0 & 4.5 & 1.0 & 0.0 \\
                & Imagination & 91.1 & 78.8 & 77.1 & 75.6 & 69.5 & 56.9 & 39.7 & 22.3 & 6.5 & 0.6 & 0.0 \\ \hline
\multirow{2}{*}{$d_{intra}$} & Baseline & 315.4 & 288.9 & 271.8 & 249.5 & 216.2 & 170.4 & 120.0 & 73.5 & 38.0 & 8.7 & 0.0 \\
                & Imagination & 104.9 & 98.7 & 99.1 & 100.4 & 99.9 & 93.6 & 79.5 & 57.9 & 26.9 & 4.1 & 0.0 \\ \hline
\end{tabular}}
\caption{Inter-class distance $d_{inter}$, intra-class distance $d_{intra}$, and separability score $J$ at different levels of ImageNet object occlusion for ``baseline'' and ``imagination'' networks.}
\label{table:LDA}
\end{table}

To illustrate this point visually, we plotted the projection of the instances of five object classes with the highest accuracy improvements on the first two principal components of the data set (Fig.~\ref{fig:analysis_PCA}). The results showed that, the ``imagination'' network maintains relatively high $J$ value with occlusion level increasing, showing its invariance to occlusion. On the contrary, the PCA result for the ``baseline'' network collapses very quickly. It indicates that the imagination process has helped to learn occlusion invariance features of these five classes, making them more discriminable under high level of occlusion. 

\subsubsection{Attention to Features}
\begin{figure}[htbp]
    \centering
    \includegraphics[width=0.7\columnwidth]{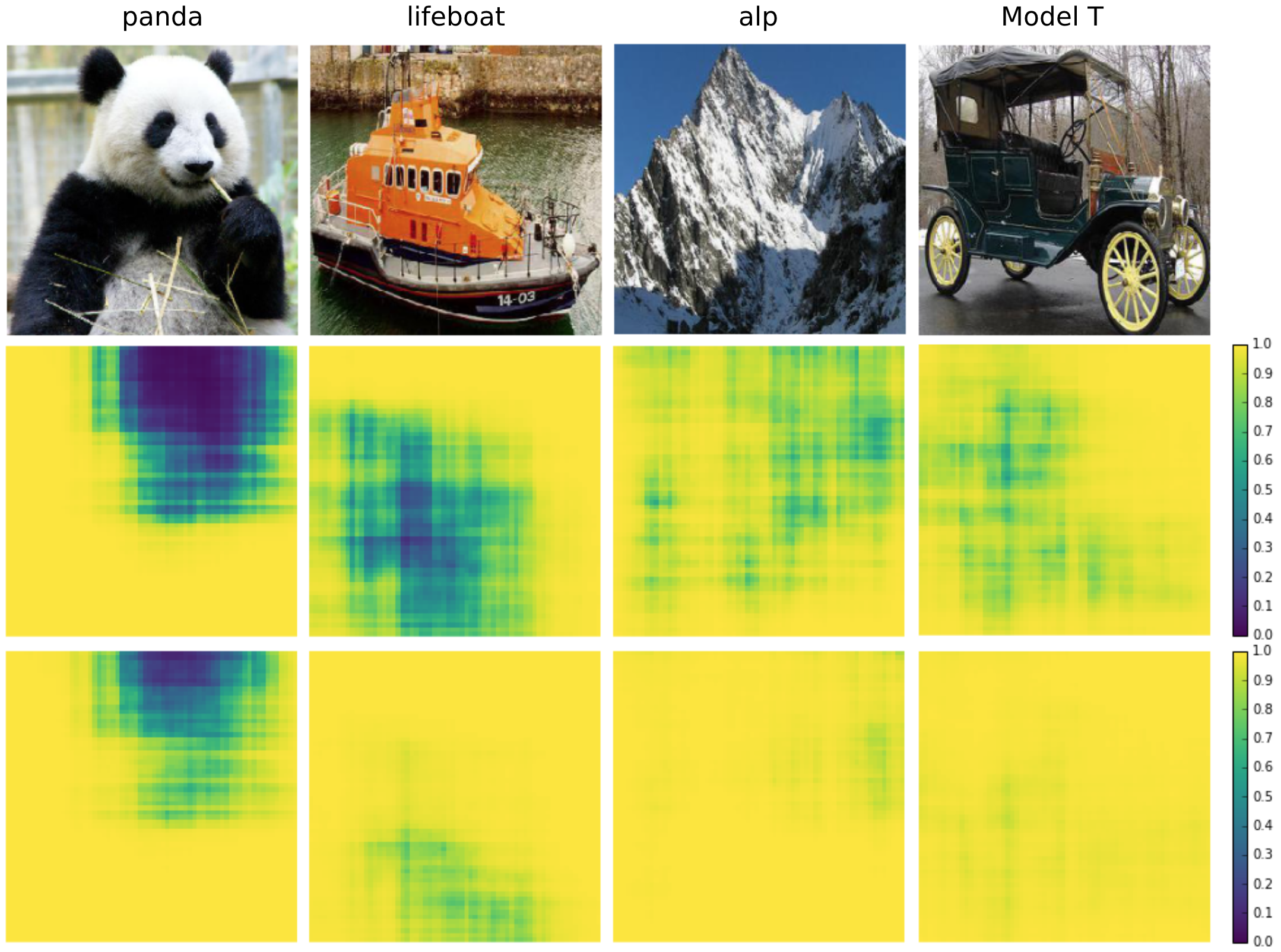}
    \caption{Confidence heat maps. First row: images for illustration. Second: heat maps from the ``baseline'' network. Third: heat maps from the ``img-rect-all'' network. Each pixel in the heat map indicates the network's confidence to target class in this position. Brighter color means higher confidence.}
    \label{fig:attention}
\end{figure}

A natural question to the phenomenon of the occlusion invariance features is the deeper mechanism behind this phenomenon. We think the ``imagination'' network achieved this partly by paying attention to more subtle and localized features in the images. Given that the ``imagination'' network learns to be robust to gray rectangular occlusion by learning from it, here we use the ``img-rect-all'' network to illustrate the idea. For both ``baseline'' and ``img-rect-all'' networks, we applied gray rectangular occluders of level 60\% in a sliding window fashion over all possible locations in each example image in the top row of Fig.~\ref{fig:attention}, and obtained the network's classification score at each location, resulting in confidence heat maps of these networks, shown in the second and third rows of Fig.~\ref{fig:attention}.

On these example images, we found that the ``img-rect-all'' network in general obtained higher confidence in larger areas. It becomes insensitive to occlusion in object parts that earned little confidence from the ``baseline'' network. Examples include the head of the panda, the cabin of the lifeboat and the cab of the Model T in Fig.~\ref{fig:attention}. Because the heat maps generated here were computed from images without full view of objects, the ``img-rect-all'' network has higher confidence over individual image parts in general.

This heat map analysis suggested that the imagination process helped the network to 1) discover and focus on some subtle and previously neglected features relevant to each class of objects; 2) not to overly rely on the presence of all parts generally associated with an object class, but more on some critical parts of the image, which is related to results mentioned in \cite{Ullman:2016ee,Wang:2015uf}.

\section{Conclusion}

In computer vision, occlusion problem is usually tackled with part-based models. For example, Deformable part model (DPM) and hierarchical DPM have been successfully applied in localizing occluded objects. However, these methods may require data sets with hand-annotated parts, and are not suitable for flexible objects or objects without parts. The success of our approach hints the possibility of automatically discovering meaningful parts of objects implicitly through an imagination composition process.

Our current work focused on evaluating the idea that data synthesized by internal models of a visual system (a neural network in our case) through composition could improve the system's ability to deal with novel situations. Our strategy could be considered as a form of data augmentation, and the case of simple occlusion (without a synthesized object as occluder) could be considered as a form of drop-out that can drive the system to discover more flexible mappings to the object recognition neurons. The most interesting insight from our work is that it will work even when the occluders are real objects, as long as the pairing of these occluding objects and the target object is rare, random and accidental, which is true in our random processes of occluder generation, similar to dreams. We found that the network can indeed discover meaningful and relevant features of local parts or atoms of recognition from these imagined or composed data to gain flexibility and robustness in recognizing objects in new contexts such as occlusion. Our results demonstrated that this is a useful idea and meaningful strategy for training deep neural networks to recognize objects under occlusion, and more generally to recognize objects in novel and unobserved situations. Whether the brain actually uses its feedback connections to implement this intriguing strategy is open to question, but the significant improvement such a strategy provides for object recognition under occlusion is interesting in its own right as a meaningful technical contribution in computer vision.

\section*{Acknowledgment}

Hao Wang and Xingyu Lin were supported by the PKU-CMU summer internship program. This work is supported by NSF (NSF grant number: NSF CISE 1320651) and by IARPA via DOI contract D16PC00007. The U.S. Government is authorized to reproduce and distribute reprints for Governmental purposes notwithstanding any copyright annotation thereon. The views and conclusions contained however should not be interpreted as necessarily representing the official policies or endorsements of the funding agencies.

\bibliographystyle{IEEEtran}
\bibliography{egbib.bib}



% \end{thebibliography}

\end{document}